\documentclass[10pt,twocolumn,letterpaper]{article}

\usepackage[pagenumbers]{cvpr} 

\definecolor{cvprblue}{rgb}{0.21,0.49,0.74}
\usepackage[pagebackref,breaklinks,colorlinks,allcolors=cvprblue]{hyperref}
\hypersetup{
 colorlinks=true,
 linkcolor=red,
 citecolor=cyan,
 filecolor=magenta, 
 urlcolor=cyan,
 }
 
\usepackage{algorithm}
\usepackage{algorithmic}
\usepackage{amsmath}
\usepackage{amssymb}
\usepackage{mathtools}
\usepackage{multirow}
\usepackage{pifont}
\newcommand{\cmark}{\ding{51}} 
\newcommand{\xmark}{\ding{55}} 

\def\eg{\emph{e.g.,~}}
\def\ie{\emph{i.e.,~}}
\def\ournet{FutureX}

\def\paperID{2603} 
\def\confName{CVPR}
\def\confYear{2026}

\title{FutureX: Enhance End-to-End Autonomous Driving \\ via Latent Chain-of-Thought World Model}

\author{
    Hongbin Lin\textsuperscript{\rm 1,2},
    Yiming Yang\textsuperscript{\rm 1,2},
    Yifan Zhang\textsuperscript{\rm 4},
    Chaoda Zheng\textsuperscript{\rm 3$\dagger$}, \\
    Jie Feng\textsuperscript{\rm 5}, Sheng Wang\textsuperscript{\rm 3}, Zhennan Wang\textsuperscript{\rm 3}, Shijia Chen\textsuperscript{\rm 3}, Boyang Wang\textsuperscript{\rm 3}\thanks{Project leader.},\\
    Yu Zhang\textsuperscript{\rm 3}, Xianming Liu\textsuperscript{\rm 3},
    Shuguang Cui\textsuperscript{\rm 2,1}, 
    Zhen Li\textsuperscript{\rm 2,1}\thanks{Corresponding authors.} \\
    \textsuperscript{\rm 1} FNii-Shenzhen,
    \textsuperscript{\rm 2} SSE, CUHK-Shenzhen,
    \textsuperscript{\rm 3} Xpeng Motors, \\
    \textsuperscript{\rm 4} MiroMind AI, 
    \textsuperscript{\rm 5} Xidian University,
}

\begin{document}
\maketitle
\begin{abstract}
In autonomous driving, end-to-end planners learn scene representations from raw sensor data and utilize them to generate a motion plan or control actions.
However, exclusive reliance on the current scene for motion planning may result in suboptimal responses in highly dynamic traffic environments where ego actions further alter the future scene.
To model the evolution of future scenes, we leverage the World Model to represent how the ego vehicle and its environment interact and change over time, which entails complex reasoning.
The Chain of Thought (CoT) offers a promising solution by forecasting a sequence of future thoughts that subsequently guide trajectory refinement.
In this paper, we propose {\textbf{\ournet}}, a CoT-driven pipeline that enhances end-to-end planners to perform complex motion planning via future scene latent reasoning and trajectory refinement.
Specifically, the Auto-think Switch examines the current scene and decides whether additional reasoning is required to yield a higher-quality motion plan. 
Once \ournet~enters the {Thinking} mode, the Latent World Model conducts a CoT-guided rollout to predict future scene representation, enabling the Summarizer Network to further refine the motion plan.
Otherwise, \ournet~operates in an \textbf{Instant} mode to generate motion plans in a forward pass for relatively simple scenes.
Extensive experiments demonstrate that \ournet~enhances existing methods by producing more rational motion plans and fewer collisions without compromising efficiency, thereby achieving substantial overall performance gains, \eg 6.2 PDMS improvement for TransFuser on NAVSIM.
Code will be released.
\end{abstract}    
\section{Introduction}
\label{sec:intro}

End-to-end (E2E) autonomous driving refers to pipelines that learn a direct and fully differentiable mapping from raw multi-modal sensor streams to a motion plan or low-level actuation commands~\cite{chen2024end}.
The field has witnessed rapid progress in both algorithmic approaches~\cite{chitta2022transfuser,hu2023planning,jiang2023vad,liao2025diffusiondrive,jia2025drivetransformer,sun2025sparsedrive} and benchmarks~\cite{caesar2020nuscenes,dosovitskiy2017carla,dauner2024navsim}.
Despite the inherent challenges, existing methods still achieve remarkable progress.

Behind the success, existing E2E autonomous driving systems directly map sensor inputs to control outputs through a single neural network, performing 
an effective one-shot feed-forward prediction without thinking further.
As a result, they struggle with adaptivity and interpretability in complex environments (2nd row in Fig.~\ref{teaser}). 
In human cognition, human drivers mentally simulate possible futures to predict how surrounding vehicles might move, how the scene may evolve, and what each possible action could lead to (top row in Fig.~\ref{teaser}) before executing any maneuver. This form of internal reasoning allows humans to make safe and context-aware decisions.
Therefore, it is essential for E2E systems to infer future scenes under highly dynamic traffic. 

Advanced large language models such as ChatGPT5~\cite{openai2025gpt5} and Qwen3~\cite{qwen3} demonstrate strong reasoning capabilities through Chain-of-Thought (CoT) mechanisms. Inspired by these developments, recent studies in autonomous driving~\cite{tian2024drivevlm,sima2024drivelm,hwang2024emma} have explored incorporating CoT-style reasoning~\cite{wei2022chain} into planning and decision-making.
Yet, these approaches primarily operate in the textual domain, generating linguistic explanations or high-level rationales that remain disconnected from the actual control process. Their ``thoughts" exist only in words, not in actions. Consequently, the resulting CoT serves more as a descriptive commentary than a functional reasoning mechanism that improves planning quality or safety. This gap motivates a new question: how can CoT reasoning be made actionable and embedded within the decision-making process itself?
We address this by reinterpreting CoT through the lens of \emph{state evolution and action selection}. We argue that the essence of CoT lies not in its textual form but in its ability to unfold the future step by step, reasoning about what will happen next and how to act accordingly. To this end, we introduce \emph{latent CoT reasoning}, where each reasoning step corresponds to a forward rollout of a latent world model followed by an internal policy evaluation. This establishes a differentiable and learnable interface between reasoning (thought) and planning (action).

Building on this insight, we propose \textbf{\ournet}, a novel end-to-end driving framework that integrates Chain-of-Thought reasoning within a Latent World Model. 
To be specific, \ournet~performs iterative ``think-simulate–act" cycles, allowing the model to reason over hypothetical futures before executing a motion.
\ournet~first introduces a Auto-think Switch, inspired by the auto-reasoning trigger of ChatGPT5~\cite{openai2025gpt5}, which decides whether to activate the world model by estimating the planning difficulty of the current scene, thereby producing a thinking or instant signal.
Then, the latent CoT reasoning is directly performed within the latent scene features based on the Latent World Model, enabling reasoning over rich spatial–temporal representations that capture environmental dynamics.
Eventually, the {Summarizer Network} predicts offsets based on the future representations together with the initial motion plan, allowing the policy network to plan on additional future information rather than exclusive reliance on the current scene.
Experiments on challenging autonomous-driving the benchmark (\eg NAVSIM~\cite{dauner2024navsim}) show that \ournet~significantly improves model performance over strong E2E baselines.

Our contributions are threefold:
1) \textbf{Conceptual}: We redefine CoT for E2E-AD as latent future reasoning—explicit state evolution and action selection within a learnable WM–policy loop.
2) \textbf{Methodological}: We propose \ournet, the first CoT-driven latent world model equipped with an Auto-Think switch that selectively activates reasoning under uncertainty, achieving a balanced trade-off between performance and efficiency for real-time deployment.
3) \textbf{Empirical}: Even with classical backbones (\ie LTF and TransFuser~\cite{chitta2022transfuser}), FutureX achieves state-of-the-art performance in both camera-only and camera–LiDAR settings, demonstrating the effectiveness and broad applicability of the proposed method.

\section{Related Work}
\label{sec:rw}

\begin{figure*}[t]
\centering
\includegraphics[width=\textwidth]{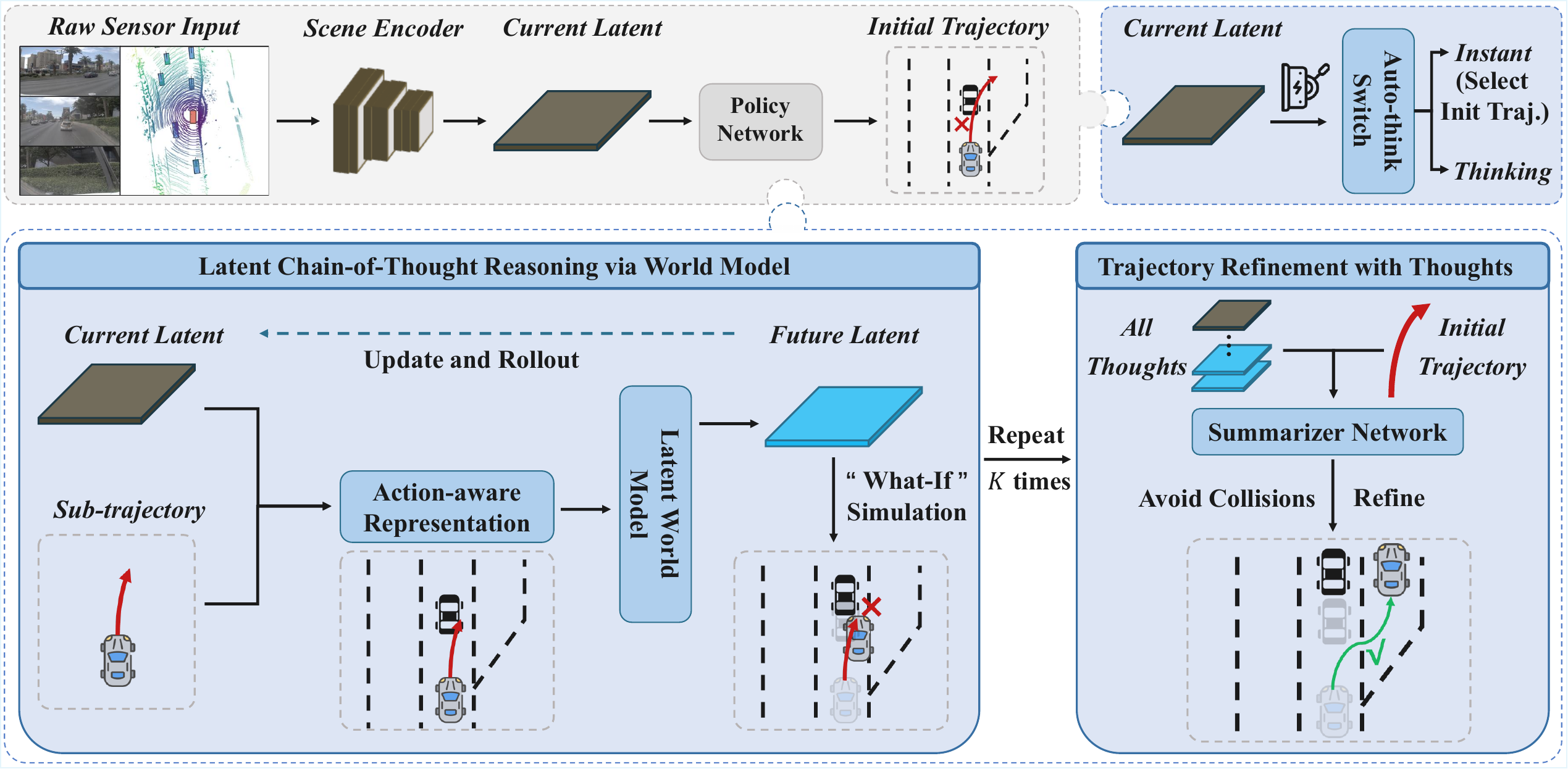}
\caption{An illustration of \ournet. 
In E2E autonomous driving, the scene encoder extracts latents from raw sensor inputs and the policy network decodes the initial trajectory. 
\ournet~leverages the Auto-think Switch to measure the difficulty of the current scene based on the latent and output an instant or thinking signal.
Once triggered, the Latent World Model performs explicit reasoning over scene latents, guiding the Summarizer Module to plan on additional future latents rather than exclusive reliance on the current scene.
}
\label{fig:method}
\end{figure*}

We first review the literature on E2E autonomous driving, and then discuss the usage of the world model in this field.
\noindent\textbf{End-to-end autonomous driving.}
The existing E2E autonomous driving methods could be divided into two categories~\cite{chen2024end}, \ie Imitation Learning (IL) and Reinforcement Learning.
On the one hand, IL methods attempt to learn from demonstrations and trains an agent to learn the policy by imitating the behavior of an expert. 
Traditional methods~\cite{pomerleau1988alvinn,bojarski2016end} generate control signals via end-to-end neural networks from the camera-only inputs, while many existing methods~\cite{prakash2021multi,chitta2022transfuser,liao2025diffusiondrive} achieve further improvements by applying multi-sensor inputs or auxiliary tasks. For the camera-only setting, the model robustness~\cite{lin2025monotta,oh2025monowad,lin2025drivegen,lin2025driveflow} of E2E methods require further explorations.
On the other hand, RL methods aim to estimate the rewards for all candidate actions in the current state and execute the action with the highest predicted value~\cite{mnih2015humanlevel}. 
Since the gradients obtained via RL are insufficient to train deep architectures for driving, several existing methods~\cite{toromanoff2020end,chekroun2023gri} adopts supervised learning with auxiliary tasks to address this issue.

\noindent\textbf{World model.}
Based on diverse inputs, world models learn real-world dynamics to capture underlying physical and spatial structure.
In autonomous driving,  there is a growing trend to conduct future state prediction for informative scene representation development.
Specifically, several existing methods~\cite{hu2023gaia,wang2024driving,russell2025gaia,zhang2025epona} aim to predict camera inputs of future scenes based on powerful diffusion models~\cite{rombach2022high,esser2024scaling,flux2024}.
However, it is time-consuming and not suitable for the deployment.
To save the cost, LAW~\cite{li2024enhancing} and World4Drive~\cite{zheng2025world4drive} utilizes latent world models to either improve the scene representation in a self-supervised learning manner or integrate driving intentions to select the suitable trajectory.
In addition, WoTE~\cite{li2025end} focuses on the trajectory evaluation side by predicting future BEV states along all candidate trajectories. 

Compared with previous methods, \ournet~firstly introduces a CoT paradigm to the latent world model, enhancing the ability of end-to-end planners to perform complex motion planning in challenging scenarios adaptively.
In addition, \ournet~automatically decides the usage of the latent world model via the auto-think switch, which achieves a balanced trade-off of performance and energy consumption. 

\section{Preliminary}
\label{sec:prel}

\noindent
\textbf{Latent World Model.}
World models aim to learn a compact predictive model of the environment dynamics from raw sensory inputs, capturing how the world evolves over time under different actions~\cite{ha2018world}. With a world model, the agent can roll out more possible future states before committing to a real action, which enables imagination-based planning.
In this work, we define a \emph{latent world model} as a differentiable transition function that models the temporal evolution of the environment \emph{within a latent feature space} that abstracts the real world. 
Rather than operating in raw observation space, the model learns to represent complex scene dynamics through structured latent states, enabling efficient and differentiable simulation of future states conditioned on ego actions.

\noindent
\textbf{Chain-of-Thought.}
Given the initial input and the desired output, Chain-of-Thought (CoT) reasoning improves complex problem solving by introducing a sequence of intermediate steps, \ie known as thoughts~\cite{wei2022chain}.
CoT is often realized as natural-language explanations, which can be viewed as a step-by-step process that bridges the input and the final answer.
However, these thoughts can also be instantiated in the action space if we construct a sequence of maneuvers or waypoints that gradually transform the current state into the desired outcome.
For instance, the low-level control commands are subsequently translated from the dynamic environment and ego-intention, \eg slow down for the pedestrian, keep the lane until passing the crosswalk, then change to the left lane.
That is why we introduce {latent CoT reasoning} for autonomous driving, where each reasoning step is realized by a forward rollout of a latent world model to simulate the dynamic environment and an internal policy evaluation that realizes the intention of the ego-vehicle in this predicted future.

\section{CoT-Driven E2E Autonomous Driving}
\label{sec:method}

\subsection{Initial Trajectory Proposal}
Without loss of generality, we denote the pipeline $f_\Theta(\cdot)$ that comprises a scene encoder to extract scene latents $\mathbf{z}_t \in \mathbb{R} ^{L \times C}$ from inputs $\mathbf{x}_t$.
Then, a policy network $\pi: \mathbb{R} ^{L \times C} \mapsto \mathbb{R} ^{T \times 3} $ predicts a full ego trajectory based on the current latent $\mathbf{z}_t$:
\begin{align}
    \mathbf{w}_t &= \pi(\mathbf{z}_t) = \{\, w_1, w_2, \dots, w_T \,\} \in \mathbb{R}^{T \times 3}.
\end{align}
Each waypoint $w_i=(x_i, y_i, \theta_i) \in \mathbb{R}^3$ is defined in the ego coordinate frame at time $t$, where $x_i$ and $y_i$ denote the spatial coordinates, and $\theta_i$ indicates the heading orientation. The whole trajectory describes the intended motion plan of the ego-vehicle over a horizon of $T$ steps.

\subsection{Latent Chain-of-Thought Reasoning}
As shown in the bottom of Figure~\ref{fig:method}, {Latent World Model} performs a CoT-guided rollout to reason future scene latents conditioned on the current latent and the initial trajectory.

\noindent
\textbf{Chain-of-Thought Segment Construction.}
To enable the structured reasoning, the trajectory $\mathbf{w}_t$ is evenly divided into $K$ sub-trajectories:
\begin{equation}
\label{eqn:waypoints}
\begin{aligned}
    \mathbf{w}_t &= [\,\mathbf{w}_t^{(1)}, \mathbf{w}_t^{(2)}, \dots, \mathbf{w}_t^{(K)}\,] \in \mathbb{R}^{K \times N \times 3}, \\
    \mathbf{w}_t^{(k)} &= \{ w_{(k-1)N+1}, \dots, w_{(k)N} \} \in \mathbb{R}^{N \times 3}.
\end{aligned}
\end{equation}
where $\mathbf{w}_t^{(k)}$ has a fixed length $N=\frac{T}{K}$. 
Each segment represents a short, local plan used for one reasoning step.



\noindent
\textbf{CoT-guided Latent World Model Rollout.}
Starting from the current latent state $\mathbf{z}_t^{(0)} = \mathbf{z}_t$, the latent world model $\mathcal{W}$ performs segment-level ``what-if" simulations, modeling how the scene evolves in latent space if the sub-trajectory is executed. 
Formally, it is formulated as:

$$
\mathcal{W}(\cdot,\cdot): \mathbb{R}^{L \times C} \times \mathbb{R}^{N \times 3} \mapsto \mathbb{R}^{L \times C},
$$
\begin{equation}
\label{eqn:fut_latent}
    \mathbf{z}_t^{(k)} = 
    \mathcal{W}\big(\mathbf{z}_t^{(k-1)},\, \mathbf{w}_t^{(k)}\big),
    \quad k = 1, \dots, K.
\end{equation}
Here $\mathcal{W}$ outputs the updated latent $\mathbf{z}_t^{(k)}$, representing the imagined latent after executing that sub-trajectory.
This process yields a sequence of future-aware latent states:
\begin{equation}
\label{eqn:thoughts}
    \mathbf{Z}_{{CoT}} =
\{\, \mathbf{z}_t^{(0)}, \mathbf{z}_t^{(1)}, \dots, \mathbf{z}_t^{(K)} \,\},
\end{equation}
which together form a latent reasoning chain—a stepwise internal simulation of the environment evolving under the ego plan.
Each reasoning step thus corresponds to a mental ``thought" about one segment of the future.
 
In our implementation, $\mathcal{W}$ is a stack of Transformer layers~\cite{vaswani2017attention}. $\mathbf{z}_t^{(k-1)}$ and $\mathbf{w}_t^{(k)}$ are first fused to form the input sequence. A trajectory encoder $E_{{traj}}(\cdot): \mathbb{R}^{N \times 3} \rightarrow \mathbb{R}^{1 \times C}$ encodes $\mathbf{w}_t^{(k)}$ into a compact embedding $\mathbf{c}_t^{(k)}$ that matches the feature dimension of $\mathbf{z}_t^{(k-1)}$.
$\mathbf{c}_t^{(k)}$ is then concatenated with $\mathbf{z}_t^{(k-1)}$ along the sequence dimension, yielding the final input sequence to the transformer layers. 
Through multi-head self-attention, the model integrates trajectory-conditioned dynamics and spatial-temporal context to produce the updated latent state $\mathbf{z}_t^{(k)}$. More details are provided in Appendix~\emph{\textbf{A}} of the supplementary material.

\subsection{Trajectory Refinement with Thoughts}
\label{sec:refine}
After generating the internal reasoning chain $\mathbf{Z}_{t}^{{CoT}}$ paired with the initial trajectory $\mathbf{w}_t$, \ournet~performs a reasoning summary step analogous to how LLMs consolidate intermediate thoughts into a final answer~\cite{guo2025deepseek,gpto1}.

As shown in the Figure~\ref{fig:method}, a summarizer network $\mathcal{S}$ takes both $\mathbf{Z}_{t}^{{CoT}}$ and $\mathbf{w}_t$ as input and predicts a refined trajectory:

$$
\mathcal{S}(\cdot,\cdot): \mathbb{R}^{K \times L \times C} \times \mathbb{R}^{K \times N \times 3} \mapsto \mathbb{R}^{T \times C},
$$
\begin{equation}
\label{eqn:refine}
\begin{aligned}
\mathbf{w}_t^{{ref}} = \mathcal{S}\big(\mathbf{Z}_{t}^{{CoT}}, \mathbf{w}_t\big),
\end{aligned}
\end{equation}
where $\mathbf{w}_t^{{ref}} = \{\tilde{w}_1, \dots, \tilde{w}_T\}$ is the refined trajectory as the final action output.

To be specific, $\mathcal{S}$ predicts the offsets of the initial trajectory $\mathbf{w}_t$ based on the predicted internal reasoning chain $\mathbf{Z}_{t}^{{CoT}}$, allowing the policy network to plan on additional future latents rather than exclusive reliance on the current latent $\mathbf{z}_t$. 
For instance, on a lane lined with parked vehicles, CoT-style reasoning helps the model anticipate pedestrians that may step out between cars in the future and thus keep a more conservative speed.

This summarization yields the final CoT-informed plan, ensuring that all internal thoughts are coherently distilled into a unified, future-consistent driving trajectory.

\subsection{Auto-think Switch}
The auto-think switch $\mathcal{G}(\cdot)$ evaluates the current scene latent $\mathbf{z}_t$ derived from raw sensor inputs, estimates the scene difficulty to determine whether to activate the latent world model $\mathcal{W}$ and output a thinking or instant signal as shown in the top of Figure~\ref{fig:method}.

Particularly, the scene encoder maps the raw sensor inputs $\mathbf{x}_t$ to a unified representation, \ie $\mathbf{z}_t$.
Conditioned on $\mathbf{z}_t$, the Auto-think Switch $\mathcal{G}(\cdot)$ outputs a scalar motion-planning difficulty score $d_t$ at timestep $t$ as below:
\begin{equation}
\label{eqn:score}
d_t=\mathcal{G}(\mathbf{z}_t), \quad d_t \in [0,1].
\end{equation}

As for the label of $d_t$, we first compute the $\ell_1$ losses between both the initial trajectory $\mathbf{w}_t$ and the refined trajectory $\mathbf{w}_t^{{ref}}$ with respect to the ground truth $\mathbf{w}_t^{{gt}}$:
\begin{equation}
\label{eqn:err}
e_{{init}} = \|\mathbf{w}_t - \mathbf{w}_t^{{gt}}\|_{1}, \quad
e_{{ref}} = \|\mathbf{w}_t^{{ref}} - \mathbf{w}_t^{{gt}}\|_{1}.
\end{equation}

Then we measure the refinement gain as the relative reduction in $\ell_1$ error and derive a binary supervision signal for the switch.
We define the calculation of the improvement ratio $r_t$ and the thinking flag $g_t$ as:
\begin{equation}
\label{eqn:g_flag}
r_t = \frac{{e}_{{init}} - e_{{ref}}}{{e}_{{init}} + \varepsilon}, \quad g_t = \mathbb{I}(r_t > \alpha), 
\end{equation}
where $\varepsilon>0$ ensures numerical stability, \(\mathbb{I}(\cdot)\) denotes the indicator function and $\alpha$ is a pre-defined threshold that modulates the sensitivity of the thinking-mode.

\subsection{Supervision in Chain-of-Thought}
To train \ournet~end-to-end, we supervise both the external trajectory predictions and the internal reasoning process.
Concretely, we define three loss terms:
1) The trajectory planning loss aligns the refined trajectories with human experts;
2) The latent consistency loss supervises the CoT latent world model;
3) The switch supervision loss teaches the Auto-think Switch when to invoke latent reasoning.

\noindent
\textbf{Latent consistency loss.}
To optimize the $\mathcal{W}(\cdot)$, we attempt to align the predicted future latents $\mathbf{z}_t^{(k)}$ with the corresponding real future latents $\hat{\mathbf{z}}_t^{(k)}$ extracted from the corresponding sensor inputs $\mathbf{{x}}_{t}$ by the scene encoder.
Therefore, we compute the latent consistency loss $\mathcal{L}_{{lat}}$ by:
\begin{equation}
\label{eqn:loss_lat}
\mathcal{L}_{{lat}} = \frac{1}{K} \sum_{k=1}^{K} \big\|\hat{\mathbf{z}}_t^{(k)} - \mathbf{z}_t^{(k)}\big\|_{1}.
\end{equation}

\noindent
\textbf{Trajectory loss.}
Since the trajectory outputs of \ournet~depends on the thinking flag $g_t$ of Eqn.(\ref{eqn:g_flag}) from auto-think switch $\mathcal{G}(\cdot)$, the final trajectory loss $\mathcal{L}_{traj}$ is computed by:
\begin{equation}
\label{eqn:traj_loss}
\mathcal{L}_{traj} = g_t \cdot e_{{ref}} + (1 - g_t) \cdot {e}_{{init}},
\end{equation}
which equips \ournet~with the ability to address difficult planning tasks through additional latent reasoning, while still preserving rapid response within relatively simple scenes.


\noindent
\textbf{Auto-Think loss.}
Given the planning difficulty score $d_t$ of Eqn.(\ref{eqn:score}) and the thinking flag $g_t$ of Eqn.(\ref{eqn:g_flag}), we calculate the cross-entropy thinking loss $\mathcal{L}_{{think}}$ by:
\begin{equation}
\label{eqn:auto_loss}
\mathcal{L}_{auto} = - [ y_t \log d_t + (1 - y_t)\log(1 - d_t) ].
\end{equation}

Overall, the training scheme of \ournet~is as follows: 
\begin{equation}
\label{loss:total}
\min_{{\Theta}} \mathcal{L}_{traj} + \lambda_1\mathcal{L}_{lat} + \lambda_2\mathcal{L}_{auto}
\end{equation}
where $\lambda_1$ and $\lambda_2$ are hyper-parameters. The pseudo-code of our training process is summarized in Algorithm~\ref{al:training} while the inference process is provided in Appendix \textbf{\emph{B}}.

\begin{algorithm}[t]
    \caption{The training pipeline of our \ournet}\label{al:training}
    \begin{algorithmic}[1]
        \REQUIRE Sensor inputs $\mathbf{x}$; Human trajectories $\mathbf{{W}}$; Model $f_{\Theta}(\cdot)$; Hyper-parameters $\alpha$, $\lambda_1$, $\lambda_2$.
    \\
    \FOR {the inputs $\mathbf{x}_t$ at the timestep $t$} 
        \STATE Extract $\mathbf{z}_t$ from $\mathbf{x}_t$ via the scene encoder;
        \STATE Decode $\mathbf{z}_t$ into $\mathbf{{w}}_t$ by the policy network;
        \STATE Prepare CoT segments by Eqn.(\ref{eqn:waypoints});
        \FOR {sub-trajectories $k = 1 \to K$}
            \STATE Calculate $\mathbf{{z}}^{(k)}_{t}$ based on Eqn.(\ref{eqn:fut_latent});
            \STATE Encodes $\mathbf{{w}}_t$ into $\mathbf{c}_t^{(k)}$ by $E_{traj}(\cdot)$;
        \ENDFOR
        \STATE Get all thoughts $\mathbf{Z}_{t}^{{CoT}}$ by Eqn.(\ref{eqn:thoughts});
        \STATE Refine $\mathbf{{w}}_{t}$ to $\mathbf{{w}}^{ref}_{t}$ by Eqn.(\ref{eqn:refine});
        \STATE Get the thinking flag $g_t$ based on Eqn.(\ref{eqn:score}), (\ref{eqn:err}), (\ref{eqn:g_flag});
        \STATE Compute latent loss $\mathcal{L}_{lat}$ based on Eqn.(\ref{eqn:loss_lat});
        \STATE Compute trajectory loss based on Eqn.(\ref{eqn:traj_loss});
        \STATE Compute think loss $\mathcal{L}_{auto}$ based on Eqn.(\ref{eqn:auto_loss});
        \STATE Get the gradient of $\Theta$ based on Eqn.(\ref{loss:total}).
        \STATE Update the model parameters of $f_{\Theta}(\cdot)$;
    \ENDFOR
    \\
    \RETURN Output the final trajectory $\mathbf{{w}}_{t}^{ref}$ if {$g_t = 1$} otherwise $\mathbf{{w}}_{t}$.
     \end{algorithmic}
\end{algorithm}

\section{Experiments}

We conduct experiments on the closed-loop NAVSIM~\cite{dauner2024navsim} dataset to evaluate the effectiveness of our approach.
In particular, we consider both a vision-only (\ie LTF) and multimodal (\ie TransFuser) E2E planners~\cite{chitta2022transfuser} as our backbone architectures.
Due to the page limitation, we further provide more experimental results in Appendix \textbf{\emph{C}}.

\noindent
\textbf{Datasets.}
The NAVSIM~\cite{dauner2024navsim} dataset is constructed on top of the nuPlan~\cite{caesar2021nuplan} benchmark. 
To be specific, OpenScene~\cite{peng2023openscene} first reduces the sampling rate from 10 Hz to 2 Hz.NAVSIM then resamples data from OpenScene to increase the prevalence of challenging scenarios.
Following the official guidelines, we divide the data into 1,192 training and 136 test scenarios, including more than 100,000 keyframes in total.

The CARLA~\cite{dosovitskiy2017carla} simulator is a widely used platform for evaluating E2E autonomous driving systems. 
Specifically, we adopt the Town1 and Town5 scenarios of the Longest6 benchmark~\cite{chitta2022transfuser} to further assess closed-loop performance,  consisting of 12 routes within 6 various weather conditions.

\begin{table*}[t]

\setlength\tabcolsep{9pt}
    \begin{center}
    \caption{\label{tab:navsim} Comparison with the E2E baselines on NAVSIM (\ie Navtest).  NC: No at-fault Collision. DAC: Drivable Area Compliance. EP: Ego Progress. TTC: Time-To-Collision. Comf.: Comfort. PDMS: Predictive Driver Model Score. C: Camera. L: LiDAR.}
        \scalebox{0.8}{
         \begin{tabular}{l|c|c|c|ccccc|cccc}
         \toprule
        Method & Input Modality & World Model & Thinking Scene & NC & DAC & EP & TTC  & Comf. & PDMS$\uparrow$ \\
        \midrule
        UniAD~\cite{hu2023planning} & C & \xmark & None & 97.8 & 91.9  & 78.8 & 92.9 & \textbf{100.0} & 83.4 \\
         VADv2~\cite{chen2024vadv2} & C & \xmark & None & 97.9 & 91.7  & 77.6 & 92.9 & \textbf{100.0}  & 83.0 \\
         LAW~\cite{li2024enhancing} & C & \cmark & All & 96.4 & 95.4  & 81.7 & 88.7  & 99.9  & 84.6 \\
         World4Drive~\cite{zheng2025world4drive} & C & \cmark & All & 97.4 & 94.3 & 79.9 & 92.8 & \textbf{100.0}  & 85.1 \\
         LTF~\cite{chitta2022transfuser} & C & \xmark & None & 97.4 & 92.8  & 79.0 & 92.4 & \textbf{100.0} & 83.8 \\
         ~$\bullet~$FutureX-Auto (Ours) & C & \cmark & Auto-select & 99.0 & 	96.3 & 		83.6 & 	95.7  & \textbf{100.0} &	{89.2}  \\
         ~$\bullet~$FutureX-All (Ours) & C & \cmark & All &  \textbf{99.6} & \textbf{96.6} &	\textbf{84.5}  & 	\textbf{96.5}  & 99.8 & \textbf{90.1} \\
         
         \midrule
         DRAMA~\cite{yuan2024drama} & C \& L & \xmark & None & 98.0 & 93.1  & 80.1 & 94.8 & \textbf{100.0}  & 85.5 \\
         Hydra-MDP~\cite{li2024hydra} & C \& L & \xmark & None & 98.3 & 96.0  & 78.7 & 94.6 & \textbf{100.0} & 86.5 \\
         DiffusionDrive~\cite{liao2025diffusiondrive} & C \& L & \xmark & None & 98.2 & 96.2 & 82.2 & 94.7 & {100.0} & 88.1 \\
         WoTE~\cite{li2025end} & C \& L & \cmark & All & 98.5 & 96.8 & 81.9 & 94.9 & \textbf{100.0}  & 88.3 \\
         TransFuser~\cite{chitta2022transfuser} & C \& L & \xmark & None & 97.7 & 92.8  & 79.2 & 92.8 & \textbf{100.0} & 84.0 \\
         ~$\bullet~$FutureX-Auto (Ours) & C \& L & \cmark & Auto-select & 99.3 & 97.1  & 84.4 & 96.2 & 99.8 & {90.2} \\
         ~$\bullet~$FutureX-All (Ours) & C \& L & \cmark & All & \textbf{99.6} & \textbf{97.2 }& \textbf{84.8} & \textbf{96.7} & 99.8 & \textbf{90.6} \\
         \bottomrule
         \end{tabular}
         }
    \end{center}
\end{table*}

\begin{table*}[t]
\setlength\tabcolsep{3pt}
    \begin{center}
    \caption{\label{tab:carla_res} Comparisons with TransFuser on the Longest6 (Town1, Town5) benchmark of CARLA.}
        \scalebox{0.8}{
         \begin{tabular}{ll|ccc|ccccccccc}
         \toprule
         Scenario & Method & DS & RC & IS & Ped & Veh & Stat & Red  & Stop & OR & Dev & TO & Block \\
         \midrule
        \multirow{2}{*}{Town1} & TransFuser~\cite{chitta2022transfuser} & 73.3  & 95.9  & 0.77  & 0.19  & 0.12  & 0.0  & 0.0  & 0.0  & 0.0  & 0.0  & 0.19 &  0.0 \\
        & ~$\bullet~$FutureX-All & \textbf{84.3}$\pm$2.2  &  \textbf{97.5}$\pm$1.5  &  \textbf{0.85}$\pm$0.03 &  0.$\pm$0. &  0.05$\pm$0.02  &  0.$\pm$0. &  0.01$\pm$0.01 &  0.$\pm$0. &  0.$\pm$0. &  0.$\pm$0. &  0.02$\pm$0.01 &  0.$\pm$0.\\
         \midrule
        \multirow{2}{*}{Town5} & TransFuser~\cite{chitta2022transfuser} &          35.1  & 69.4  & 0.63  & 0.0  & 1.38  & 0.0  & 0.0  & 1.38  & 0.09  & 0.0  & 0.17  & 0.52 \\
        & ~$\bullet~$FutureX-All & \textbf{53.7}$\pm$1.2  &  \textbf{87.1}$\pm$4.8  &  \textbf{0.63}$\pm$0.05 &  0.$\pm$0. &  0.07$\pm$0.03 &  0.$\pm$0. &  0.$\pm$0. &  0.03$\pm$0.01 &  0.$\pm$0. &  0.$\pm$0. &  0.02$\pm$0.01 & 0.02 \\            
         \bottomrule
         \end{tabular}
         }
    \end{center}
\end{table*}

\noindent\textbf{Implementation details.}
We implement our method and all baselines in PyTorch~\cite{paszke2019pytorch}, and train every model on 16 NVIDIA H20 GPUs for a total of 50 epochs.
We employ the Adam optimizer~\cite{adam2014method} with a learning rate of 1e-4 and a batch size of 128.
The $\alpha$, $K$, $\lambda_1$ and $\lambda_2$ are set to $0.25$, $2$, $0.1$, $0.1$ (\ie $N=2$), respectively. 
FutureX-Auto is our default setting while FutureX-All activates the think mode for all scenes.
More details are provided in Appendix~\textbf{\emph{D}}.

\noindent\textbf{Compared methods.}
Built on the LTF or TransFuser backbone~\cite{chitta2022transfuser}, we compare the proposed \ournet~with the following state-of-the-art (SOTA) E2E driving methods:
(1) \emph{Vision-only baselines}: LTF~\cite{chitta2022transfuser}, UniAD~\cite{hu2023planning}, and VADv2~\cite{chen2024vadv2};
(2) \emph{Vision-only baselines with world models}: LAW~\cite{li2024enhancing} and World4Drive~\cite{zheng2025world4drive};
(3) \emph{Multimodal baselines}: TransFuser~\cite{chitta2022transfuser}, DRAMA~\cite{yuan2024drama}, and Hydra-MDP~\cite{li2024hydra};
(4) \emph{Multimodal baselines with world models}: WoTE~\cite{li2025end}.

\noindent\textbf{Evaluation protocols.}
We evaluate model performance using the closed-loop PDM Score (PDMS), which aggregates five terms: no at-fault collisions (NC), drivable area compliance (DAC), time-to-collision (TTC), comfort (Comf.), and ego progress (EP).
The PDMS is computed by:
$$PDMS=NC\times DAC\times \frac{5\times EP + 5\times TTC + 2Comf.}{12}.$$

As for CARLA, we evaluate model performance via Driving Score (DS) computed from Route Completion (RC) and Infraction Score (IS). RC measures the fraction of each route completed by the agent, whereas IS summarizes the total number of infractions.
Besides, the remaining infractions-per-kilometer metrics are defined as follows: Ped, collisions with pedestrians; Veh, collisions with vehicles; Stat, collisions with static layout; Red, red-light violations; OR, off-road driving; Dev, route deviation; TO, timeout; and Block, vehicle blocked.
For our method, we report the mean and standard deviation over three runs.

\subsection{Comparisons with Previous Methods}
We first analyze the comparisons between our \ournet~and previous methods. The results presented in Table~\ref{tab:navsim}
gives the following observations:
1) In vision-only and multimodal settings, recent methods such as World4Drive~\cite{zheng2025world4drive} and WoTE~\cite{li2025end} achieve substantial performance improvements over the classical baselines like UniAD~\cite{hu2023planning} and TransFuser~\cite{chitta2022transfuser} by integrating latent world models for future scene representation development;
2) Building on the classical LTF and TransFuser backbone, \ournet-Auto~still achieves consistently performance improvements on all metrics except Comfort (which is very close to the maximum), achieving 5.4 and 6.2 PDMS gains for LTF and TransFuser, respectively; 
3) Even in the auto-think setting, \ournet-Auto~achieves the current SOTA performance by only performing latent CoT reasoning and trajectory refinement with thoughts on a subset instead of all scenes;
4) For fair comparisons, we report the performance of \ournet-All which conducts thinking in all scenarios.
Remarkably, \ournet-All achieves \emph{over 90 PDMS} on top of LTF, and further demonstrates strong performance by reaching \emph{90.6} PDMS with TransFuser.

As for the Longest6 benchmark, Table~\ref{tab:carla_res} shows that \ournet~also achieves comprehensive improvements in terms of a total of 12 metrics.
To be specific, the base TransFuser gets \emph{\textbf{11.0}} and \emph{\textbf{18.6}} driving score gains in Town1 and Town5 with \ournet, which reflects the overall capability of the model and serves as the primary criterion for comparison.
Hence, these results further verify our \ournet.


\begin{figure*}[t] 
  \centering
  \includegraphics[width=0.8\linewidth]{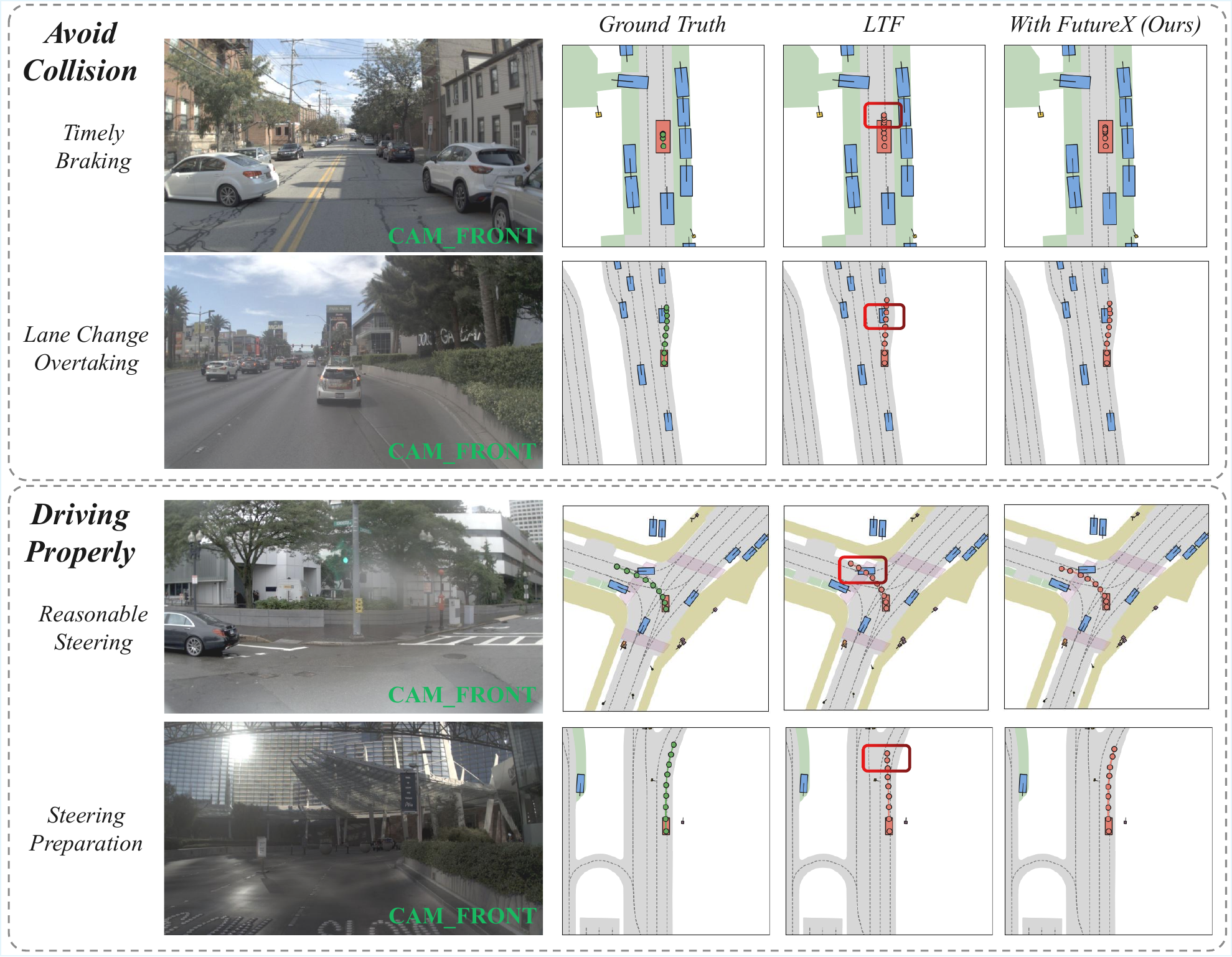} 
   \caption{Qualitative results of \ournet~which avoids collisions by timely braking or lane change overtaking (top), and offers a proper driving plan with more reasonable steering (bottom).
   }
   \label{fig:our_vis}
   
\end{figure*}

\subsection{Beyond Performance Gains}
\label{exp:analysis}
An intuitive concern arises regarding how the final motion plan is improved after integrating \ournet.
To investigate this, we leverage the official visualization tool to inspect several representative cases enhanced by \ournet as shown in Figure~\ref{fig:our_vis}.
Interestingly, we observe that \ournet~enables the planner to revise its initially predicted trajectories which would lead to collisions or suboptimal maneuvers.

In the first two scenarios, LTF~\cite{chitta2022transfuser} produces the initial trajectories that pose a potential collision risk.
Specifically, 
1) the first scenario shows the front vehicle is reversing on a narrow road. With \ournet~integrated, the model brakes in time and successfully avoids the collision.
2) The second row presents an overtaking case in which the baseline planner generates an insufficient lane-change action. With \ournet, the model identifies this potential issue and produces a more appropriate lane-change trajectory.

In addition, sharp turns at intersections pose a significant challenge for autonomous driving systems. As observed in robotaxi, their systems may fail to generate a feasible steering maneuver, causing emergency stop in the intersection.
3) Here, the third row illustrates that \ournet~rectifies the under-steering issue of the initial trajectory and
4) the fourth row further demonstrates \ournet~is capable of anticipating future scenes and preparing actions ahead of time.
By reasoning hypothetical futures before execution, \ournet~is able to anticipate the risks induced by the original driving behavior, and subsequently refine its actions to ensure safe and reliable navigation.
 
\begin{figure*}[t] 
  \centering
  \includegraphics[width=0.8\linewidth]{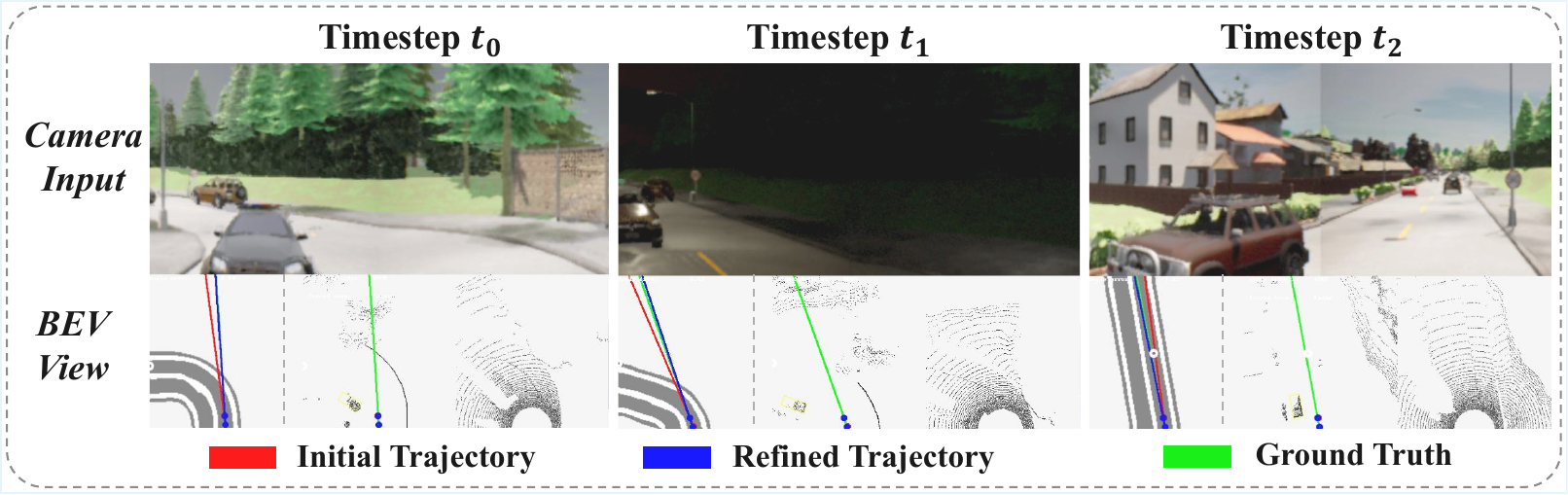} 
   \caption{Qualitative results of trajectory refinement of \ournet~on CARLA. More visualizations are put in Appendix \textbf{\emph{E}}.
   }
   \label{fig:carla}
\end{figure*}

\begin{table*}[t]
\setlength\tabcolsep{9pt}
    \begin{center}
    \caption{\label{tab:abla_cot} Ablation studies on the length $N$ of sub-trajectories with different values. Length $T$ of predictive trajectories $\mathbf{w}_t$ is 8 in NAVSIM.}
        \scalebox{0.8}{
         \begin{tabular}{l|c|c|c|cccc|ccccc}
         \toprule
        Method  & CoT & $K$ trajectories & Length $N$ & NC $\uparrow$ & DAC $\uparrow$ & EP $\uparrow$ & TTC $\uparrow$ & PDMS$\uparrow$ \\
        \midrule
         LTF~\cite{chitta2022transfuser} &  \xmark & 1 & 8  & 97.4 & 92.8  & 79.0 & 92.4  & 83.8 \\
         ~$\bullet~$FutureX-Auto & \cmark & 2 & 4 & 98.6 & 95.5 & 82.4 & 94.8  & 87.8   \\
         ~$\bullet~$FutureX-Auto & \cmark & 4 & 2  & 99.0 & 96.3 & 83.6 & 95.7 & \textbf{89.2}  \\
         \bottomrule
         \end{tabular}
         }
    \end{center}
\end{table*}

\begin{table}[t]
\setlength\tabcolsep{4pt}
    \begin{center}
    \caption{\label{tab:abla_loss} Ablation studies on the training loss terms
     $\mathcal{L}_{lat}$ and $\mathcal{L}_{traj}$ of \ournet~based on LTF~\cite{chitta2022transfuser}, which are also the ablation studies of \emph{latent CoT reasoning} and \emph{trajectory refinement with thoughts}. 
     }
        \scalebox{0.8}{
         \begin{tabular}{ccc|cccc|cccccc}
         \toprule
        Backbone & $\mathcal{L}_{lat}$ & $\mathcal{L}_{traj}$ & NC $\uparrow$ & DAC $\uparrow$ & EP $\uparrow$ & TTC $\uparrow$ & PDMS$\uparrow$  \\
        \midrule
         \cmark & \xmark & \xmark & 97.4 & 92.8  & 79.0 & 92.4 & 83.8 \\
         \cmark & \cmark & \xmark & 99.7  & 	95.1 & 	82.8 & 	97.0 & 	88.7 (\textcolor{red}{+4.9$\uparrow$}) \\
         \cmark & \cmark & \cmark & 99.6 & 96.6 & 84.5  & 96.5 & \textbf{90.1} (\textcolor{red}{+6.3$\uparrow$}) \\
         \bottomrule
         \end{tabular}
         }
    \end{center}
\end{table}

\begin{table}[t]
\setlength\tabcolsep{9pt}
    \begin{center}
    \caption{\label{tab:abla_latency} Ablation studies on the latency of \ournet~with different $N$, which is measured on a single NVIDIA H20 GPU.}
        \scalebox{0.8}{
         \begin{tabular}{l|ccccccc|ccccc}
        \toprule
        Methods & Length $N$ &  Latency (ms) & PDMS \\
        \midrule
        LTF~\cite{chitta2022transfuser} & $N=8$ & 2.3 $\pm$ 0.1 & 83.8 \\
        ~$\bullet~$FutureX-All & $N=4$ & 17.0 $\pm$ 0.3 & 87.8 \\
        ~$\bullet~$FutureX-All & $N=2$ & 31.3 $\pm$ 0.6 & 89.2 \\
        \bottomrule
         \end{tabular}
         }
    \end{center}
\end{table}

\subsection{Ablation Studies}
\noindent\textbf{Effectiveness of loss components.}
To examine each component in \ournet, we first present quantitative results of different loss settings based on LTF~\cite{chitta2022transfuser} as shown in Table~\ref{tab:abla_loss}.
For fair comparisons, we adopt the thinking mode for all scenes.
Compared with the backbone, introducing the latent consistency loss $\mathcal{L}_{lat}$ activates the latent chain-of-thought reasoning. Thus, the model achieves more informative scene representations, thereby achieving a comprehensive performance improvement.
Next, by summarizing the predicted internal reasoning chain, the trajectory loss $\mathcal{L}_{traj}$ mainly improves the model with higher drivable area compliance (DAC), and ego progress (EP). 
As illustrated in Section~\ref{sec:refine}, the policy network is able to plan on additional future latents (\ie $\mathbf{Z}_{t}^{{CoT}}$) rather than exclusive reliance on the current latent $\mathbf{z}_t$, getting more reasonable motion plans.
As for the auto-think loss $\mathcal{L}_{auto}$, we can see the comparisons between \ournet-Auto and \ournet-All within Table~\ref{tab:navsim}.
Even though we adopt the instant mode for 24.5\% and 13.7\% scenes, the performance drop of \ournet~are 0.9 and 0.4 PDMS for LTF and TransFuser, which still maintains state-of-the-art performance.
Notably, even the instant trajectory benefits from the optimization of the scene representations.
This result further demonstrates the effectiveness and practical value of the proposed method.

\noindent\textbf{Chain-of-Thought Reasoning.}
As shown in Table~\ref{tab:abla_cot}, we conduct additional ablation studies on the length of sub-trajectories $N$.
To be specific, the default length $T$ of predictive trajectories $\mathbf{w}_t$ is set to 8 in NAVSIM.
Thus, we divide $\mathbf{w}_t$ into 2 or 4 sub-trajectories for explorations.
Intuitively, with more reasoning steps, \ournet~achieves the better performances in terms of 5 metrics since the planner accesses more hypothetical futures and conducts more iterative ``think-simulate–act" cycles.

\noindent\textbf{Analysis of latency.}
A common concern is the total latency of latent CoT reasoning and trajectory refinement of our method.
For this concern, we further provide the ablation studies on the latency of \ournet~based on a single NVIDIA H20 GPU.
As shown in Table~\ref{tab:abla_latency}, it only requires 17.0 ms when the length of sub-trajectories $N=4$, which still outperforms all baselines.
As the length $N$ decreases, \ournet~achieves stable gains while only increasing the latency up to 31.3 ms, which still satisfies the demands for real-time
deployment in autonomous driving.

   

\subsection{Quantitative Results}
We provide qualitative visualizations of \ournet~as shown in Figure~\ref{fig:our_vis} and \ref{fig:carla}.
As mentioned in Section~\ref{exp:analysis}, the Figure~\ref{fig:our_vis} shows the relatively complex scenes with the Thinking mode of \ournet, which improves the motion plan.
Besides, we also provide the qualitative results of CARLA.
Figure~\ref{fig:carla} reveals that \ournet~improves the initial motion
and avoids the collision with the opposite black car.


\section{Conclusion}
\label{sec:conclu}
In this paper, we propose an end-to-end driving framework integrates chain-of-thought reasoning within a latent World Model, namely \ournet.
Our method attempts to reason over hypothetical futures before executing the action by performing iterative ``think-simulate–act" cycles.
Specifically, the latent world model conducts latent CoT reasoning over rich scene representations, thereby capturing environmental dynamics. 
Then \ournet~performs a summary of reasoning step analogous to LLMs consolidating thoughts into a final answer, which yields the final CoT-informed plan.
For practical deployment, an auto-think switch decides whether to activate the latent world model, achieving a balanced trade-off of performance and time consumption.
Experiments on the NAVSIM and CARLA demonstrate the effectiveness and applicability of \ournet~in enhancing both vision-only and multimodal end-to-end autonomous driving.



{
    \small
    \bibliographystyle{ieeenat_fullname}
    \bibliography{main}
}


\end{document}